\documentclass[onecolumn]{article}

\makeatletter

\newenvironment{affiliations}{%
    \setcounter{enumi}{1}%
    \setlength{\parindent}{0in}%
    \slshape\sloppy
    \begin{list}{\upshape$^{\arabic{enumi}}$}{%
        \usecounter{enumi}%
        \setlength{\leftmargin}{0in}%
        \setlength{\topsep}{0in}%
        \setlength{\labelsep}{0in}%
        \setlength{\labelwidth}{0in}%
        \setlength{\listparindent}{0in}%
        \setlength{\itemsep}{0ex}%
        \setlength{\parsep}{0in}%
        }
    }{\end{list}\par\vspace{10pt}}

\renewenvironment{abstract}{%
    \noindent\textbf{Abstract}---\setlength{\parindent}{0in}%
    \setlength{\parskip}{0in}%
    }{\par\vspace{-5pt}}
\makeatother

\usepackage{tabulary,graphicx,amsmath,amsfonts,amssymb,setspace}
\usepackage[figurename=Fig.]{caption}
\usepackage{booktabs}
\usepackage{tabularx}
\usepackage{multirow}
\usepackage{float}
\usepackage{subfig}
\usepackage{graphicx}
\usepackage[margin=0.7in]{geometry}
\usepackage{hyperref}
\usepackage{lettrine}
\usepackage{hyperref}
\usepackage[dvipsnames,table]{xcolor}
\usepackage[sort&compress,numbers]{natbib}
\usepackage{lineno}
\usepackage{color}

\usepackage{url}

\hypersetup{
bookmarks=true,
bookmarksopen=true,
bookmarksnumbered=true,
unicode=false,
pdftoolbar=true,
pdfmenubar=true,
pdffitwindow=false,
pdfstartview={FitH},
pdftitle={My title},
pdfauthor={Author},
pdfsubject={Subject},
pdfcreator={Creator},
pdfproducer={Producer},
pdfkeywords={keywords},
pdfnewwindow=true,
colorlinks=true,
linkcolor=blue,
citecolor=blue,
filecolor=blue,
urlcolor=blue
}

\title{\bf Multimodal foundation models are better simulators\\ of the human brain}

\author{ 
    Haoyu Lu\(^{1,2,}\)\thanks{These authors contributed equally to this work.}~,
    Qiongyi Zhou\(^{3,4,*}\),
    Nanyi Fei\(^{2,5,*}\),
    Zhiwu Lu\(^{1,2}\),
    Mingyu Ding\(^{6}\),
    Jingyuan Wen\(^{1,2}\),\\
    Changde Du\(^{3,4}\),
    Xin Zhao\(^{1,2}\),
    Hao Sun\(^{1,2}\),
    Huiguang He\(^{3,4}\),
    Ji-Rong Wen\(^{1,2,7}\) 
}

\date{}

\begin{document}

\maketitle

\vspace{-0.1in}
\begin{affiliations}
  \item 
    Gaoling School of Artificial Intelligence\unskip, 
    Renmin University of China\unskip, Beijing 100872, China
  \item 
    Beijing Key Laboratory of Big Data Management and Analysis Methods\unskip,
    Beijing 100872, China
  \item 
    Research Center for Brain-inspired Intelligence\unskip, National Laboratory of Pattern Recognition\unskip, Institute of Automation\unskip, Chinese Academy of Sciences\unskip, Beijing 100190, China
  \item 
    School of Artificial Intelligence\unskip, University of Chinese Academy of Sciences\unskip, Beijing 100049, China
  \item 
    School of Information\unskip, 
    Renmin University of China\unskip, Beijing 100872, China
  \item
    The University of Hong Kong\unskip, Pokfulam, Hong Kong
  \item
    Beijing Academy of Artificial Intelligence\unskip, Beijing, China
\end{affiliations}
\vspace{0.1in}

\begin{abstract}
Multimodal learning, especially large-scale multimodal pre-training, has developed rapidly over the past few years and led to the greatest advances in artificial intelligence (AI). Despite its effectiveness, understanding the underlying mechanism of multimodal pre-training models still remains a grand challenge. Revealing the explainability of such models is likely to enable breakthroughs of novel learning paradigms in the AI field. To this end, given the multimodal nature of the human brain, we propose to explore the explainability of multimodal learning models with the aid of non-invasive brain imaging technologies such as functional magnetic resonance imaging (fMRI). Concretely, we first present a newly-designed multimodal foundation model pre-trained on 15 million image-text pairs, which has shown strong multimodal understanding and generalization abilities in a variety of cognitive downstream tasks. Further, from the perspective of neural encoding (based on our foundation model), we find that both visual and lingual encoders trained multimodally are more brain-like compared with unimodal ones. Particularly, we identify a number of brain regions where multimodally-trained encoders demonstrate better neural encoding performance. This is consistent with the findings in existing studies on exploring brain multi-sensory integration. Therefore, we believe that multimodal foundation models are more suitable tools for neuroscientists to study the multimodal signal processing mechanisms in the human brain. Our findings also demonstrate the potential of multimodal foundation models as ideal computational simulators to promote both AI-for-brain and brain-for-AI research.
\end{abstract}

\section*{Introduction}

Although artificial intelligence (AI) has been flourishing over the past decade, critical attention has been primarily placed on developing unimodal learning models for solo tasks commonly seen in computer vision, natural language processing, etc. There have been efforts placed on multimodal learning~\cite{atrey2010multimodal, rasiwasia2010new}; however, its development was rather slow due to the limitations of cross-modal interaction techniques, training data annotations, and computational resources. Either specific approaches were devised for each modality followed by a simple fusion module, or the developed multimodal models are only suitable for a few cognitive tasks. This is clearly against the common concept that a unified model is needed to acquire multimodal recognition, understanding, or even generation abilities, since the human brain has the multimodal nature \cite{nadler1993multisensory, tan2021bioinspired,smith2005development}. Until recently, unified multimodal learning is made possible by the emergence of large-scale multimodal foundation models like CLIP~\cite{radford2021learning} and ALIGN~\cite{jia2021scaling}, which has become a hot research topic. A number of recent works have explored different designs of neural network architectures as well as pre-training objectives, leading to promising results in a variety of multimodal tasks including cross-modal retrieval~\cite{lin2014microsoft, plummer2015flickr30k, lu2022cots} and visual question answering~\cite{LeLV020, YangMSLS21}. Particularly, even on many unimodal tasks, deploying multimodal models achieves significant improvements compared with unimodal ones~\cite{radford2021learning, jia2021scaling}, which is intuitively sound because the human brain also functions better with multimodal information \cite{smith2005development,botta2011multisensory,matusz2017multisensory}. Moreover, for human brain's higher-level cognitive abilities like imagination (text-to-image generation), models with impressive semantic understanding and generation abilities have been developed \cite{abs-2204-06125, abs-2205-11487}, showing the general applicability of multimodal learning. Overall, we believe that multimodal learning is a more potential approach to cognitive AI (or even artificial general intelligence).

Although multimodal foundation models~\cite{radford2021learning, jia2021scaling, lu2022cots} have shown various cognitive abilities across a wide range of multimodal tasks, what they have actually learned (or how they have managed to possess these abilities) remains a mystery. Revealing the basic mechanism of multimodal models will deepen our understanding of why and how the model works, clear people's concerns and doubts about the models, and enable new designs of explainable learning paradigms. There are two pioneering works~\cite{goh2021multimodal, fei2022towards} making attempts to explore the explainability of multimodal foundation models: the former~\cite{goh2021multimodal} focuses only on the visual encoder of CLIP~\cite{radford2021learning} and discovers that multimodal neurons exist in this visual encoder trained in a multimodal way; the latter~\cite{fei2022towards} qualitatively presents the unified image-text embedding space. Nevertheless, they explain large-scale multimodal foundation models only in the AI research field itself, and the current understanding of multimodal models is still insufficient.

\begin{figure}[t]
\centering
\includegraphics[width=0.98\textwidth]{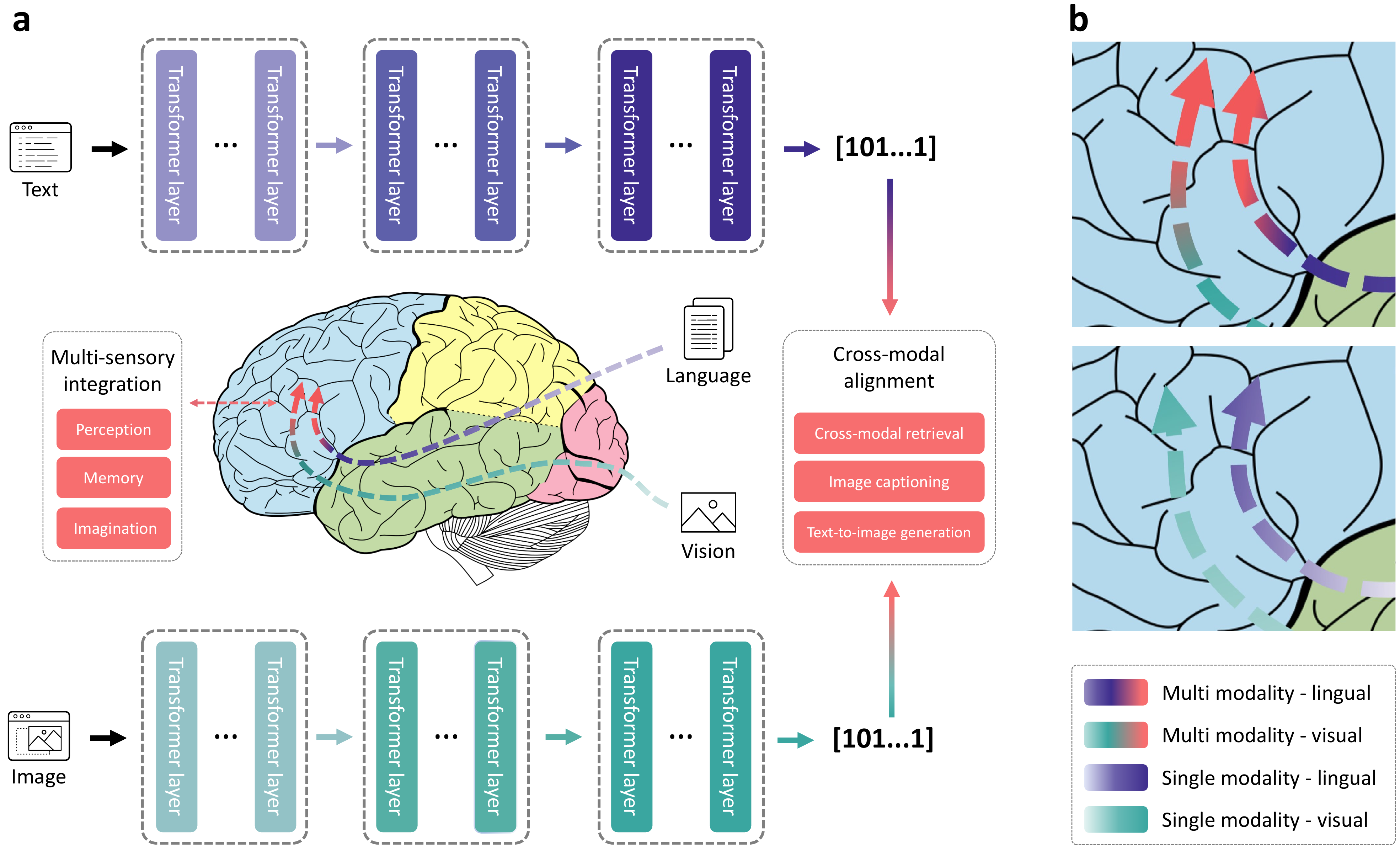}
\vspace{-0.0in}
\caption{\textbf{Overarching concept of our BriVL model.} \textbf{a}. Comparison between human brain and our multimodal foundation model BriVL for coping with both vision and language information. \textbf{b}.~Schematic illustration of processing single modality and multimodal modality in human brain. The simultaneous input of paired visual and lingual signals may activate closer (multimodal) regions in brain (converging along the red arrows), while unpaired/individual signals of different modalities may end up a bit more separately (along the arrows in different colors).}
\label{fig:fig1} 
\vspace{-0.1in}
\end{figure}

Note that the human brain is commonly found to be multimodal (see Fig.~\ref{fig:fig1}a), which receives multi-sensory information (e.g., vision and language) via different streams and then integrates them to achieve cognitive abilities (e.g., perception, memory, and imagination).
This paradigm is very similar to that of multimodal foundation models, which learn from large-scale multimodal data and thus have the generalization ability across various brain-like downstream tasks (e.g., cross-modal retrieval analogous to understanding and perception, and cross-modal generation analogous to imagination).
More importantly, paired multimodal stimuli are expected to activate multimodal brain regions to a greater extent than unimodal stimuli~\cite{goolkasian2005bimodal, saults2007central, delogu2009semantic}. 
That is, the simultaneous input of paired visual and lingual signals may activate closer (multimodal) regions in the brain, while unpaired/individual signals of different modalities may end up a bit more separately, which is schematically shown in Fig.~\ref{fig:fig1}b. This is again very similar to the core idea of multimodal foundation models~\cite{radford2021learning, jia2021scaling, lu2022cots} pre-trained with contrastive learning~\cite{he2020momentum, oord2018representation}. 
In this work, we thus propose to exploit the computational neuroscience (CN) methods for a closer look at multimodal foundation models. To this end, we choose to explore the multimodal model's explainability with the help of the widely-used non-invasive brain imaging technology -- functional magnetic resonance imaging (fMRI)~\cite{matthews2004functional}. Unlike most existing works that conduct experiments with unimodally-trained neural networks to analyze their brain-like properties~\cite{xu2021limits, yamins2016using, gucclu2015deep, ratan2021computational, zhuang2021unsupervised, jain2018incorporating, toneva2019interpreting, antonello2021low}, we focus on revealing whether multimodal models are more brain-like than unimodal ones. In this work, we only take vision and language modalities into account, as they are the most commonly studied in both the AI and CN research fields.

Concretely, we train a large-scale multimodal foundation model named Bridging-Vision-and-Language (BriVL) with a total of 15 million image-text pairs. We follow the design of dual-stream neural network architecture~\cite{radford2021learning, jia2021scaling}, where visual and lingual encoders are separate and the modality alignment is only performed at the final global embedding level (see Fig.~\ref{fig:fig1}a). This design is selected because multimodal foundation models with single- and hybrid-stream architectures~\cite{chen2020uniter, li2021align} fuse data from different modalities in a fine-grained way, making them very difficult to explain with computational neuroscience. For training our BriVL model, by extending the unimodal contrastive learning algorithm MoCo~\cite{he2020momentum}, we adopt the momentum mechanism and maintain large negative sample queues for cross-modal contrastive learning (i.e., cross-modal MoCo) with low GPU memory occupation. Once trained with cross-modal MoCo over 15 million image-text pairs, our BriVL model is shown to acquire strong cross-modal understanding and generation abilities in cross-modal retrieval and generation tasks. Furthermore, we also evaluate the neural encoding~\cite{naselaris2011encoding, nunez2019voxelwise} performance of our BriVL by making it predict the neural responses in the human brain triggered by visual/lingual stimuli. For comprehensive investigation, we compare the visual encoder of our BriVL with the same-sized unimodal Vision Transformer (ViT)~\cite{dosovitskiy2021an} on a neural dataset containing fMRI responses from five human subjects viewing movie clips~\cite{huth2012continuous, huthgallant}, and similarly compare the lingual encoder of our BriVL with the language model BERT~\cite{devlin2018bert} on a neural dataset containing fMRI signals with lingual stimuli~\cite{pereira2018toward}.

Extensive experiments on neural encoding show that both multimodally-trained visual and lingual encoders are more brain-like than their same-sized unimodally-trained counterparts. This finding indicates that multimodal foundation models are more suitable for building the computational models of the human brain than unimodal ones. Meanwhile, this also suggests that the superior performance of multimodal foundation models on various cognitive tasks is mainly due to their brain-like properties (i.e., developing brain-like models is vital for AI research). More importantly, we identify a number of brain regions (e.g., pSTS, Temporal\_Sup and Frontal\_Sup) where multimodally-trained encoders explain the neural activities better than unimodal ones. This supports the findings in existing studies on exploring brain multi-sensory integration. Therefore, multimodal foundation models are indeed more suitable tools for neuroscientists to study the multimodal signal processing mechanisms in the human brain. Overall, with deeper exploration on the correspondence between multimodal foundation models and the human brain, we believe that multimodal foundation models have the potential to serve as ideal computational simulators that promote both the AI-for-brain and brain-for-AI research.

\section*{Results}

We first conduct experiments to show that our BriVL model trained with large-scale multimodal data possesses strong cognitive capabilities, including multimodal perception and memory by cross-modal retrieval as well as content description and imagination by cross-modal generation. Moreover, with neural encoding, we investigate whether the visual and lingual encoders of our BriVL are better at encoding real neural signals in the human brain than their corresponding same-sized unimodally-trained ones. We discuss each part in detail in this section.

\begin{figure}[t]
\centering
\includegraphics[width=0.96\textwidth]{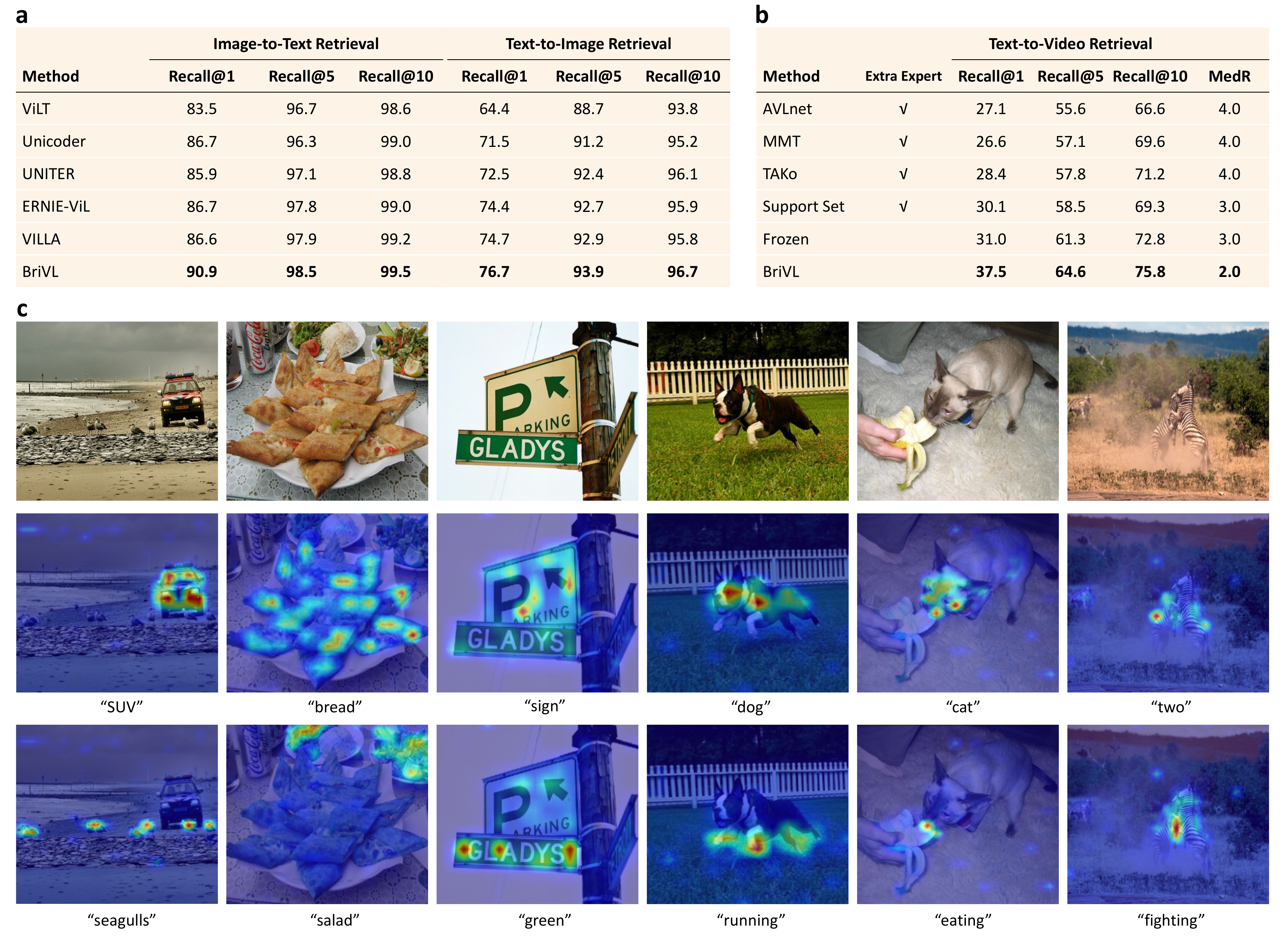}
\vspace{-0.12in}
\caption{\textbf{Cross-modal retrieval results.} ~\textbf{a.}~Image-text retrieval results (\%) on the Flick30K dataset.  ~\textbf{b.}~Text-to-video retrieval results on the MSR-VTT 1k-A dataset. Results on Recall (\%) and MedR (Median Rank) are reported.  ~\textbf{c.}~Visualizations of attention maps of our pre-trained BriVL on images responding to individual words.}
\label{fig:fig2}
\vspace{-0.12in}
\end{figure}

\paragraph{Cross-modal retrieval.}
Humans have the ability to determine whether concepts from different sensory modalities are semantically aligned. To show that our BriVL model also has such ability, we conduct experiments on cross-modal retrieval tasks. The first dataset is Flick30k~\cite{plummer2015flickr30k}, which consists of 31,000 images and 158,915 captions in total (with about 5 captions per image). Following UNITER~\cite{chen2020uniter}, we adopt the Karpath split~\cite{anderson2018bottom} which takes 30,000 images for training and the rest 1,000 images for testing. We first apply BriVL's visual/lingual encoder to obtain image/text embeddings, and then calculate the cosine similarity of each image-text pair as its similarity score. The image-text retrieval results on Flick30k are shown in Fig.~\ref{fig:fig2}a, where we report Recall@$k$\footnote{Recall@$k$ is defined as the proportion of matched samples found in the top-$k$ retrieved results.} ($k = 1, 5, 10$) following ERNIE-ViL~\cite{yu2021ernie}.
Our BriVL is compared to strong baselines which adopt single-stream network architectures designed for image-text retrieval (i.e., Unicoder~\cite{li2020unicoder} and ViLT ~\cite{kim2021vilt}) or deploy object detectors for obtaining better performance (i.e., UNITER~\cite{chen2020uniter} and VILLA~\cite{gan2020large}). It can be observed that, even without considering single-stream architectures or deploying object detectors, our BriVL still outperforms all competitors by a large margin on all metrics, showing the strong multimodal understanding ability of our pre-trained BriVL model.

To further demonstrate the general applicability of our BriVL, we conduct experiments on the text-to-video retrieval task. Compared with image-text retrieval, text-to-video retrieval is more challenging as the video modality introduces extra temporal information and contains more noise.
We evaluate our BriVL on a widely-used benchmark MSR-VTT~\cite{xu2016msr}, which contains 10,000 videos and 200,000 captions totally (each video is annotated with 20 captions). Following the standard split setting~\cite{bain2021frozen}, we take 9,000 videos for training and 1,000 videos for testing. We compare our BriVL with strong baselines (including AVLnet~\cite{rouditchenko2020avlnet}, MMT~\cite{Gabeur0AS20}, TAKo~\cite{yang2021taco}, and Support Set~\cite{patrick2020support}, and Frozen~\cite{bain2021frozen}), and report results with Recall@$k$ ($k = 1, 5, 10$) as well as MedR (i.e., the median rank of the target truth in the retrieval results) in Fig.~\ref{fig:fig2}b. Since our BriVL is pre-trained on pure image-text datasets, we directly average all frame embeddings of each video to obtain the video embedding. We find that our BriVL outperforms all competitors by large margins on all metrics, including those considering `extra expert' features~\cite{rouditchenko2020avlnet, Gabeur0AS20, patrick2020support, yang2021taco} (e.g., object, motion, face, scene, speech, and sound features). This further validates the strong general applicability of our BriVL.

Furthermore, we deploy the Generic Attention-model Explainability (GAE) technique~\cite{chefer2021generic} to visualize the attention of our pre-trained BriVL towards images responding to individual words in Fig.~\ref{fig:fig2}c. It can be seen that: (1) Our BriVL has the ability to well locate different objects (e.g., ``seagulls'', ``salad'' and ``sign'' in Fig.~\ref{fig:fig2}c). (2) Our BriVL can also capture color concepts (e.g., ``green'' in Fig.~\ref{fig:fig2}c) and actions (e.g., ``running'', ``eating'' and ``fighting'' in Fig.~\ref{fig:fig2}c). Overall, these visualization results demonstrate that our BriVL is able to identify multiple objects and concepts without introducing any fine-grained cross-modal fusion module like the single-stream network architecture.

\begin{figure*}[t]
\centering
\includegraphics[width=0.96\textwidth]{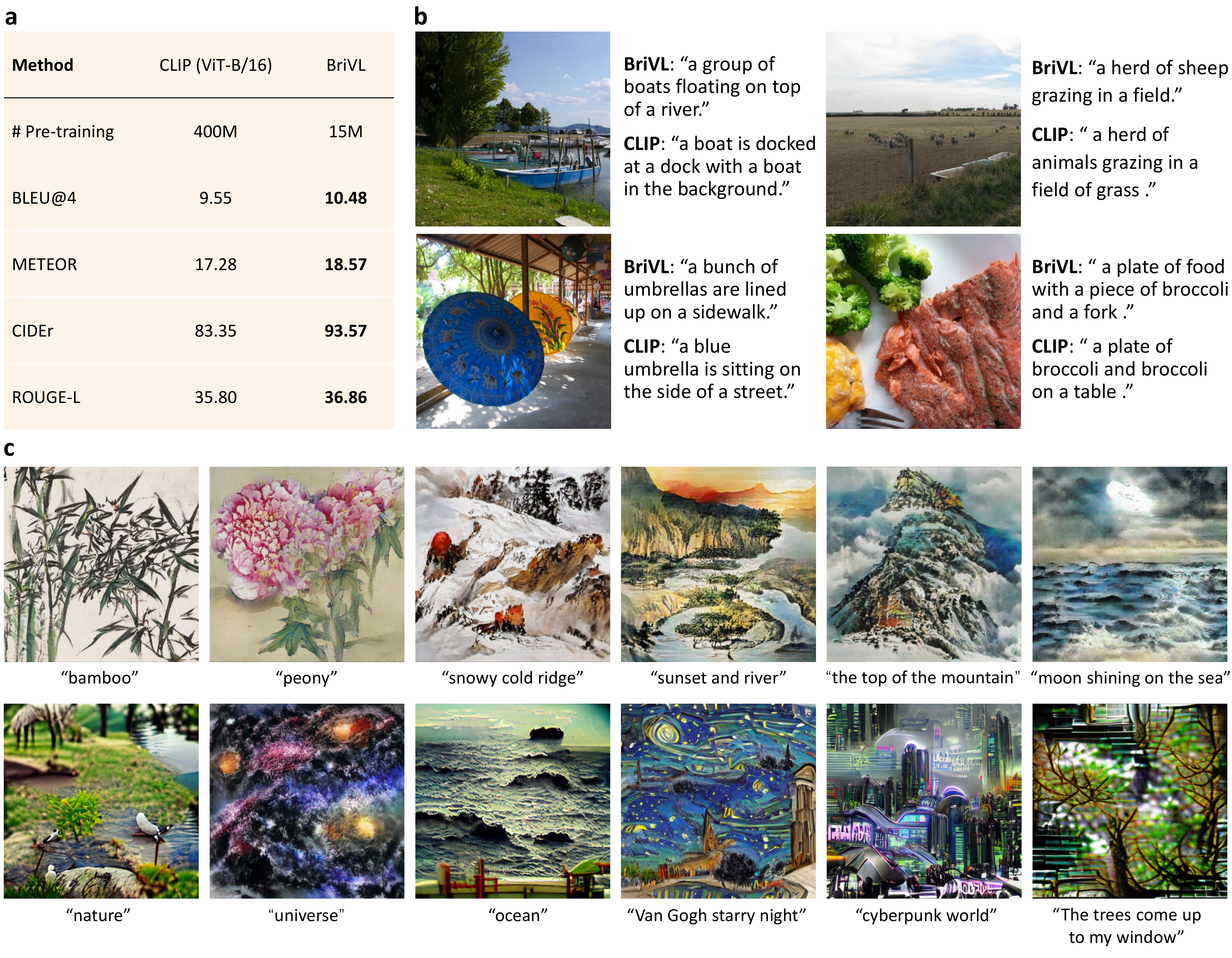}
\vspace{-0.1in}
\caption{\textbf{Cross-modal generation results.}
\textbf{a.} Image-to-text generation results on MSCOCO compared with CLIP (ViT-B/16)~\cite{radford2021learning}. \textbf{b.} Image-to-text generation examples of CLIP and our BriVL. \textbf{c.}~Text-to-image generation examples by VQGAN~\cite{esser2020vqgan} inversion with our BriVL.}
\label{fig:fig3}
\vspace{-0.1in}
\end{figure*}

\paragraph{Cross-modal generation.}
When receiving object information from one sensory modality, human can naturally imagine what the object is like in other modalities. For example, the image of a stream will come up into our minds when we hear the sound of trickling water. Similarly, our BriVL can also be applied to such cross-modal generation. To show this ability of our BriVL, we conduct two groups of experiments on bi-directional cross-modal generation tasks: image-to-text generation (i.e., image captioning) and text-to-image generation.

For image-to-text generation, we evaluate our BriVL on the MSCOCO~\cite{lin2014microsoft} dataset and compare it with CLIP~\cite{radford2021learning}. For fair comparison, only CLIP (ViT-B/16) is deployed, where ViT denotes the well-known Vision Transformer~\cite{dosovitskiy2021an}. We take 113,287 images (with 5 captions per image) for training and 5,000 images for testing. For each image, we first obtain its image embedding through the visual encoder of BriVL/CLIP. Then the same Transformer-based decoder is employed for both BriVL and CLIP to transform image embeddings into texts. We report the results on MSCOCO in the table of Fig.~\ref{fig:fig3}a with four commonly-used metrics (i.e., BLEU\@4, METEOR, CIDEr, and ROUGE-L) in image captioning. Although CLIP is pre-trained on a much larger dataset with 400M image-text pairs, our BriVL still outperforms CLIP on all four metrics, showing its effectiveness in image-to-text generation. We further provide four generation examples of BriVL/CLIP in Fig.~\ref{fig:fig3}b. It can be seen that both BriVL and CLIP can capture the main content of the given images (e.g., ``boat'' and ``umbrella'' in the left two images). However, our BriVL is more accurate in semantic understanding during image captioning than CLIP (e.g., ``sheep'' vs. ``animals'' in the upper-right image, ``a bunch of umbrellas'' vs. ``a blue umbrella'' in the lower-left image). 

Compared with image captioning, text-to-image generation is more challenging as images have higher information density. Nevertheless, our BriVL can still generate precise and artistic images as shown in Fig.~\ref{fig:fig3}c. Specifically, our BriVL can accurately generate images given object words (e.g., ``universe'' and ``peony'') and also artistic images in the landscape painting style (e.g., ``snowy cold ridge'' and ``sunset and river''). 
Our BriVL is also capable of other art painting styles (e.g., ``Van Gogh starry night'' and ``cyberpunk world''). 
Moreover, our BriVL can even generate images from long sentences with complex information (e.g., the last example in Fig.~\ref{fig:fig3}c). These examples validate the strong imagination ability of our BriVL, just like the imagination ability of the human brain.

\paragraph{Neural encoding with visual stimuli.}
We evaluate the performance of BriVL in predicting the neural responses triggered by visual stimuli. Concretely, we compare the neural encoding performance of BriVL's visual encoder with that of the same-sized unimodally-trained ViT~\cite{dosovitskiy2021an}. The neural dataset is collected by \cite{huth2012continuous, huthgallant} which contains the Blood Oxygenation Level-Dependent (BOLD) signals recorded from five subjects viewing natural movie clips. BriVL and ViT are compared layer-by-layer with the banded ridge regression model \cite{nunez2019voxelwise}. For each layer, features from BriVL and ViT are extracted and concatenated. The hemodynamic delay is also considered as in \cite{huth2012continuous}, that is, the response at second $i$ is predicted by the concatenated features at seconds $i$, $i-4$, $i-6$, and $i-8$.
Before model fitting, we select voxels with explainable variance (EV) above 0.1. We choose this threshold because the selected voxels have high-reliability and voxels outside the visual cortex can be excluded. Models are then fitted based on the neural data of each subject and the average coefficient of determination ($R^2$) of all selected voxels on all subjects is reported.

\begin{figure}[t!]
\centering
\includegraphics[width=0.98\textwidth]{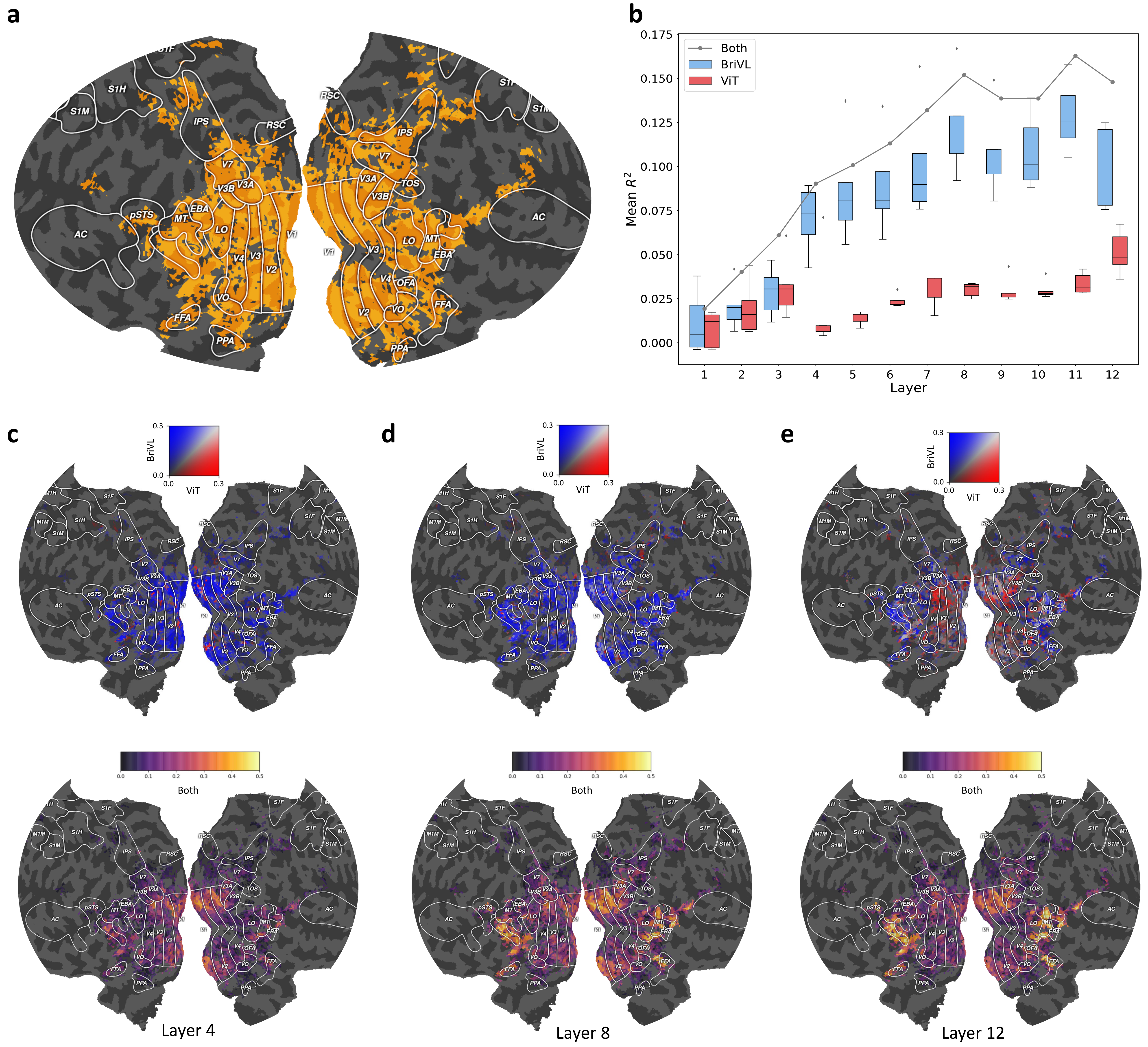}
\vspace{-0.1in}
\caption{\textbf{Neural encoding with visual stimuli.}
\textbf{a.} Locations of the selected voxels on the cerebral cortex. \textbf{b.} The mean ($R^2$) of all selected voxels on all subjects as a function of the number of network layers. \textbf{c.}~Visualizations of the encoding accuracy on Layer 4, 8, and 12 plotted on the cortical flatmap of Subject 1. Due to the limited space, we crop the visualization subfigures in \textbf{a, c--e} and thus the brain areas are slightly different.
}
\label{fig:fig4} 
\vspace{-0.1in}
\end{figure}

We present the results of visual neural encoding in Fig.~\ref{fig:fig4}. For easier reference, Fig.~\ref{fig:fig4}a shows the locations of the selected voxels on the cerebral cortex, which cover regions of interest (ROIs) including the early visual cortex (e.g., V1--V4), the high-level inferior temporal cortex (e.g., LO, EBA, FFA, and PPA, which respectively response to object, body, face, and scene), and the visual cortex in the dorsal stream (e.g., MT and V7). 
Fig.~\ref{fig:fig4}b shows the prediction accuracy ($R^2$) as a function of each network layer. The grey line plots the mean $R^2$ on all subjects using the concatenated features of BriVL and ViT (i.e., the joint encoding model, denoted as ``Both'') with the banded ridge regression. The box plots in blue and red respectively present the split $R^2$ of BriVL and ViT on five subjects. It can be seen that as the layer goes deeper, the encoding accuracy achieved by the joint encoding model has an obvious increasing tendency in the beginning and gradually becomes stable after the 8-th layer. It reflects that features from deeper model layers can enhance the overall encoding performance on the visual cortex. 
More importantly, as Fig.~\ref{fig:fig4}b shows, the contribution of BriVL is much higher than that of ViT from the 4-th layer, indicating that visual representations of these layers learn richer multimodal information and these representations are more similar to human brain responses. 

We further provide more visualization results of visual neural encoding on Layer 4, 8, and 12 in Fig.~\ref{fig:fig4}c--e, respectively. The bottom subfigures show the $R^2$ obtained by the joint encoding model (of BriVL and ViT), and the top subfigures show the split $R^2$ of BriVL and ViT, plotted on the cortical flatmap of Subject 1.
The 1-D colormap in each bottom subfigure marks the color corresponding to the $R^2$ of the joint encoding model. Brain areas with higher prediction are shown in brighter yellow. It can be seen that features of Layer 8 predict the best on the low-level visual regions, such as V1-V4, and features of Layer 12 predict the best on the high-level visual regions, such as FFA and EBA in the inferior temporal cortex and MT in the dorsal stream. This result is consistent with the prior study revealing the hierarchical correspondence between computer vision models and human visual pathway \cite{conwell2021can}. 
Meanwhile, the 2-D colormap in each top subfigure marks the color given a pair of split $R^2$ of BriVL and ViT. Concretely, in the cortical flatmap, BriVL contributes more than ViT in blue areas while ViT contributes more than BriVL in red areas. 
From the top subfigures in Fig.~\ref{fig:fig4}c--e, we can observe that the common areas where BriVL shows superiority across all three layers are the high-level visual cortex. This can be interpreted in two ways.
On one hand, several regions in the high-level visual cortex have been implicated to play an important role in multi-sensory integration~\cite{beauchamp2004unraveling, calvert2004multisensory}. Particularly, it is shown in the top subfigures in Fig.~\ref{fig:fig4}c--e that BriVL outperforms ViT in predicting the neural activities of posterior superior temporal sulcus (pSTS, a region that has been studied as an audiovisual multimodal ROI~\cite{beauchamp2004unraveling}), which demonstrates the necessity of deploying multimodal information in the neural encoding for multimodal brain regions.
On the other hand, cross-modal interactions during the multimodal model training can facilitate better representation learning of abstract semantics~\cite{lu2022cots}. Therefore, high-level visual cortices that focus on characterizing high-level semantic features can benefit from the neural encoding based on multimodal models. Both of these explanations demonstrate the intrinsic advantage of those models with multimodal learning paradigms in neural encoding.
Furthermore, we can also observe the overwhelming advantage of BriVL in the low-level visual regions from the top subfigures in Fig.~\ref{fig:fig4}c--d, which is supported by previous studies~\cite{ghazanfar2006neocortex, macaluso2000modulation, murray2016multisensory} that have shown functional and anatomical evidence that not only higher-level visual cortices but also primary visual ones participate in multi-sensory processing. Note that this advantage is not observed in Layer 12, which could be explained by the mismatch between the basic visual representations in V1-V4 and the over-abstract representations in Layer 12 of BriVL. To make the findings on visual neural encoding more convincing, visualization results of visual neural encoding on more subjects and more layers are provided in the Supplementary Fig.~S5--S13.

\paragraph{Neural encoding with lingual stimuli.}
In addition to visual neural encoding, we also evaluate the neural encoding performance of BriVL with lingual stimuli. For fair comparison, we set the counterpart of BriVL's lingual encoder to be the unimodally-trained language model BERT~\cite{devlin2018bert}.
The neural data used for lingual neural encoding is a subset of the benchmark dataset collected by ~\cite{pereira2018toward}, including three experiments.
In this work, we choose the data from Experiment 2 containing fMRI of eight subjects with 384 sentences as stimuli, since our BriVL is actually pre-trained with image-sentence pairs. Voxels are split into 116 ROIs according to the Automated Anatomical Labelling (AAL) altlas~\cite{tzourio2002automated}. Since the public data do not contain trial repetitions, we conduct voxel selection not by calculating their EV but by their location and encoding accuracy. First, we discard voxels in the last 26 ROIs in the cerebellum to focus on the cerebrum. Second, we only keep voxels with positive $R^2$ for further analyses, since it is meaningless to compare the contributions of two models on voxels with the encoding accuracy $R^2 \leq 0$.
The lingual neural encoding is still performed with the banded ridge regression~\cite{nunez2019voxelwise}, as in the neural encoding with visual stimuli. For comprehensive study, more results on Experiment 3 data are also provided in the Supplementary Fig.~S1.

\begin{figure}[t]
\centering
\includegraphics[width=0.98\textwidth]{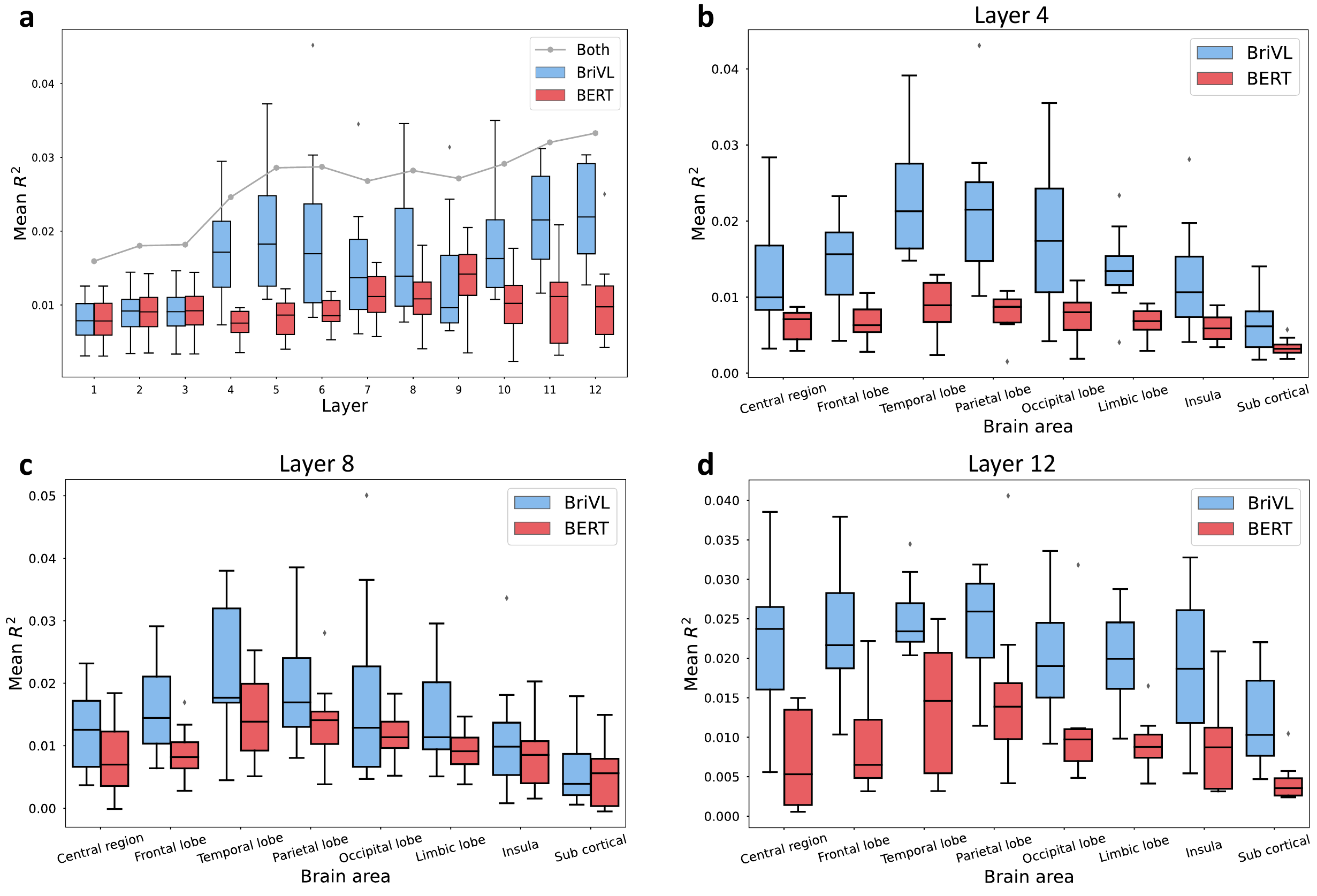}
\vspace{-0.1in}
\caption{\textbf{Neural encoding with lingual stimuli.}
\textbf{a.} The average ($R^2$) of all selected voxels on all subjects as a function of the number of network layers. \textbf{b--d.}~The average ($R^2$) of all selected voxels on all subjects with features of Layer 4, 8, and 12, as a function of different brain areas.}
\label{fig:fig5}
\vspace{-0.1in}
\end{figure}

Fig.~\ref{fig:fig5} shows the results of lingual neural encoding. The encoding performance of each model layer is presented in Fig.~\ref{fig:fig5}a. The grey line shows the prediction accuracy of the concatenated features of BriVL and BERT (i.e., the joint encoding model, denoted as ``Both'') on all subjects when the number of layers grows up. It can be seen that the explanatory power of the joint encoding model shows an increasing trend as the layer goes deeper. The box plots in blue and red respectively present the split $R^2$ of BriVL and BERT on all subjects. Since BriVL and BERT share network parameters for the first three layers, the values of the split $R^2$ of these layers of BriVL and BERT are almost equal. From the 4-th layer, the median of BriVL is higher than that of BERT except for the 9-th layer. The results show that the encoding performance of BriVL's lingual encoder is superior to that of the unimodal lingual encoder, i.e., BriVL with multimodal knowledge is closer to the neural representation of the human brain in processing linguistic information than the unimodal model. Particularly, the special case at the 9-th layer can be explained by different patterns of variation in the explanatory power of different models with layer depth. Note that the maximum explanatory power of BERT is achieved at the 9-th layer, while the maximum explanatory power of BriVL is achieved at the last layer. Therefore, the encoding accuracy of BriVL is squeezed to some degree at the 9-th layer. 
To further demonstrate the advantage of BriVL in predicting neural activities of different brain regions, we present the encoding performance of Layer 4, 8, and 12 on different ROIs in Fig.~\ref{fig:fig5}b--d, respectively. According to the previous study~\cite{tzourio2002automated}, we merge the 90 ROIs in the AAL atlas into eight big brain areas, i.e., central region, frontal lobe, temporal lobe, parietal lobe, occipital lobe, limbic lobe, insula, and sub cortical. We find that BriVL outperforms BERT over almost all regions on Layer 4, 8, and 12 with various explanatory power.
In addition, we also have a closer look at the advantage of BriVL over individual brain areas on Layer 12 in Fig.~\ref{fig:fig5}d. Note that the temporal lobe is investigated as a region involved in language processing, and the high-level brain regions like the central region, frontal lobe, and parietal lobe are known to participate in various forms of advanced cognition and perception (e.g., reasoning and memory of the frontal lobe~\cite{chayer2001frontal}, sensory perception and integration of the parietal lobe~\cite{COSLETT2018365}). From Fig.~\ref{fig:fig5}d, we can observe that our BriVL achieves higher encoding accuracy on both of the temporal lobe and these high-level brain regions, reflecting that the incorporation of multimodal information can induce better brain-like properties. Due to the limited space, we provide the detailed results of lingual neural encoding on each of eight big brain areas as a function of the number of network layers in the Supplementary Fig.~S4.

\begin{figure}[t!]
\centering
\includegraphics[width=0.98\textwidth]{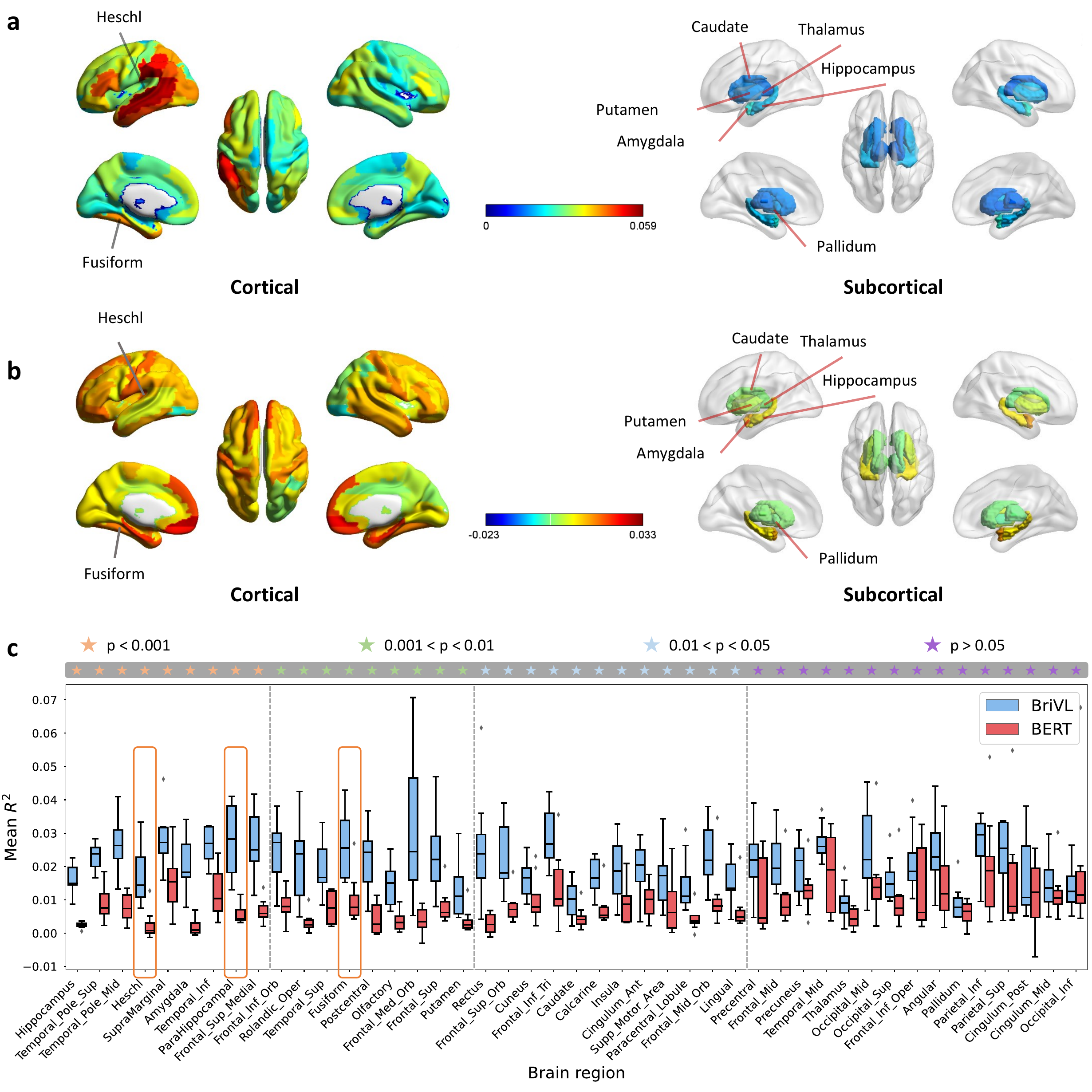}
\vspace{-0.1in}
\caption{\textbf{Neural encoding with lingual stimuli on Layer 12.}
\textbf{a.} Visualizations of the encoding accuracy of the joint encoding model.
\textbf{b.} Visualizations of the encoding accuracy difference between BriVL and BERT.
\textbf{c.} 
The average ($R^2$) of all selected voxels on all subjects, as a function of 45 brain regions.}
\label{fig:fig6}
\vspace{-0.1in}
\end{figure}

Fig.~\ref{fig:fig6} shows the detailed results of the lingual neural encoding on Layer 12. We project the prediction accuracy of the joint encoding model and the difference between the split $R^2$ of BriVL and BERT onto the MNI template and show them in Fig.~\ref{fig:fig6}a--b, respectively. The visualization tool we used is a MATLAB toolbox called the BrainNet Viewer~ \cite{xia2013brainnet}. 78 of the 90 ROIs are located in the cortex and the remaining 12 in the subcortex (i.e., the left and right hemispherical Amygdala, Putamen, Caudate, Thalamus, Hippocampus, and Pallidum), which are respectively shown in the left and right subfigures in Fig.~\ref{fig:fig6}a--b. It can be seen from Fig.~\ref{fig:fig6}a that our joint encoding model better predicts the neural activities in the left-hemisphere temporal, inferior parietal, and orbitofrontal regions (i.e., regions in dark red in Fig.~\ref{fig:fig6}a) than other regions. These well-predicted regions are consistent with the left-lateralized cortical regions of the core language network that is summarized in~\cite{hertrich2020margins}. We can also observe some regions with relatively high prediction accuracy beyond the core language network, for example the whole frontal lobe and the occipital lobe.
Recent studies~\cite{hertrich2020margins,rolls2022human} have pointed out that language processing is not only dependent on the core language network but also requires the involvement of brain areas related to motor, emotion, social cognition, etc. Our encoding results provide consistent conclusions from a computational neuroscience perspective, based on the assumption that if the variability of neural responses in a particular brain region is largely independent of the stimuli, the model cannot predict the responses well~\cite{schoppe2016measuring}.
Fig.~\ref{fig:fig6}b presents the contribution difference of BriVL and BERT. BriVL makes more contribution than BERT does on regions in warmer color and BERT makes more contribution than BriVL does on the regions in colder color. It can be seen that BriVL predicts neural activities better than BERT does in the high-order brain regions and regions of the left-hemisphere language network. This result suggests that multi-sensory integration widely exists in the language network and the high-order brain regions. Besides, advantages of BriVL are widely observed in a large scale of regions, including subcortical regions. In addition to these visualization results on Layer 12, the neural encoding performance on all layers projected onto the MNI template is provided in the Supplementary Fig.~S2 and Fig.~S3.
To further compare the different performance of BriVL and BERT, we show the encoding accuracy on 45 ROIs that are created by merging voxels in the corresponding regions in the left and right hemispheres in Fig.~\ref{fig:fig6}c. Paired two-tailed t-test is conducted on the encoding accuracy of BriVL and BERT. Along the horizontal axis, p-values of these ROIs are in ascending order. It can be seen that BriVL significantly surpasses BERT on most brain regions. 
Among these brain regions with smallest p-values, a number of brain regions have been well studied to be multimodal. For example, previous studies show that the superior temporal gyrus (Temporal\_Pole\_Sup, Temporal\_Sup) is involved in integrating audiovisual information~\cite{ozker2018converging} and is important for fluent reading~\cite{ye2017audiovisual}, and studies in~\cite{visser2012both,davey2016exploring} show that the middle temporal gyrus (Temporal\_Pole\_Mid) plays a important role in multimodal semantic processing. The supramarginal gyrus (SupraMarginal) has also been investigated to work with the angular gyrus (Angular) to transfer visual words to specific meanings~\cite{stoeckel2009supramarginal}, which is a multimodal processing. The advantage of BriVL over these brain regions suggests that multimodal pre-training models can facilitate the neural response prediction in multimodal brain regions.
Furthermore, brain regions that are considered to be associated with high-level cognitive functions also achieve smallest p-values, such as hippocampus for memory~\cite{henke1999human}, amygdala for emotional processing and memory~\cite{zald2003human}, and superior frontal gyrus (Frontal\_Sup\_Medial, Frontal\_Med\_Orb, Frontal\_Sup, and Frontal\_Sup\_Orb) for attention, execution, and working memory~\cite{li2013subregions}. This reflects that these cognitive functions may involve the integration and understanding of multimodal information, and suggests that multimodal pre-training models can learn more abstract and high-level representations than unimodal ones, resulting in better predictive performance in higher-order brain regions.
Importantly, we find that BriVL even has a significant advantage over BERT on putatively sensory-specific brain regions, such as the heschl gyrus (Heschl) which is described as the primary auditory cortex and is shown to be unimodal in~\cite{khosla2021cortical}. However, studies on primates have shown that parts of the primary auditory cortex have connections to visual cortex~\cite{budinger2006multisensory, kayser2009multisensory, ghazanfar2006neocortex}, and our results (orange boxed) support these findings. Besides, our results (orange boxed) are also consistent with the findings in the current studies on the multimodal properties in parahippocampal cortex (ParaHippocampal)~\cite{wang2017domain, wolbers2011modality} and fusiform gyrus (Fusiform)~\cite{reich2011ventral, striem2012reading}. In this way, our BriVL can be used as auxiliary analysis tool for neuroscientists to study brain functions.

\section*{Discussion}

We have developed a newly-designed multimodal foundation model named BriVL, which is pre-trained on 15 million image-text pairs. We have validated its effectiveness on downstream multimodal tasks including cross-modal retrieval and cross-modal generation. For cross-modal retrieval, BriVL achieves promising performance on image-to-text, text-to-image, and text-to-video retrieval tasks. We have also qualitatively demonstrated its semantic understanding ability by visualizing BriVL's attention on image regions w.r.t. individual words. For cross-modal generation, BriVL shows strong image captioning and text-to-image generation (imagination) abilities, which are more high-level and brain-like. More importantly, we have designed extensive experiments based on non-invasive brain imaging and found that both visual and lingual encoders trained in the multimodal way are respectively more brain-like than their same-sized unimodally-trained counterparts. This finding is of great importance because it partly has explained why multimodal foundation models possess various cognitive capabilities.

We believe that our current work on explaining multimodal foundation models with non-invasive brain imaging would have a significant impact on both the brain-for-AI and also AI-for-brain research. On one hand, with our observations, we think that future AI models could follow more brain-like designs which may lead to superior performance in various cognitive tasks. On the other hand, considering that multimodal foundation models perform better than unimodal ones in explaining signals from multimodal brain regions, carefully-designed multimodal models can be used as auxiliary analysis tools for neuroscientists to study the multimodal signal processing mechanisms in the human brain. For example, multimodal models can be deployed to explore whether a part of a brain region is multimodal, to determine the multimodal function of certain neurons/synapses, or to investigate the degree of multimodal fusion across different brain regions.
In addition, for real-world applications, we could deploy multimodal foundation models (e.g., generate images with them in a controllable way) to activate particular neurons of the human brain in the ongoing research, making a step to AI-based treatment of brain-related disease.

Nevertheless, deploying multimodal foundation models and exploring their explainability with computational neuroscience still face large challenges and high risks. Firstly, researchers should be well aware that the pre-training data used for model training may contain prejudices, which would result in unjust models and bring negative effects. Secondly, given that the human brain receives and processes information from multiple sensory modalities, most existing multimodal foundation models only considering vision and language are far from being well-developed. Including more modalities such as video and audio would be a challenge but also a must for obtaining more advanced multimodal foundation models. Thirdly, although our BriVL has been shown to have certain brain-like properties, it is not even close to completely understand the human brain and better brain-inspired models need to be developed. Last but not least, revealing the explainability of multimodal foundation models in this work is only a start. The explainability should be further explored with more advanced computational neuroscience technologies or for more complex multimodal models (e.g., with single- and hybrid-stream architectures).

\section*{Materials and Methods}

\paragraph{Pre-training data collection.}
The pre-training data collection used for training our BriVL model consists of six image-text datasets: Conceptual Captions12M~\cite{changpinyo2021conceptual}, Conceptual Captions3M~\cite{sharma2018conceptual}, SBU~\cite{ordonez2011im2text}, VG~\cite{krishna2017visual}, MSCOCO~\cite{lin2014microsoft}, and Flickr30k~\cite{plummer2015flickr30k}, which accumulate 15.2M image-text pairs in total.

\paragraph{Network architecture and pre-training objective.}
As we have stated in the Introduction section, the network architectures of existing large-scale multimodal foundation models can be mainly categorized into three classes: single-stream, dual-stream, and hybrid-stream.
Although single- and hybrid-stream models have achieved promising results, they are very difficult to pre-train with large-scale noisy data and have the disadvantage of high latency during inference. Moreover, their fine-grained cross-modal fusion modules fuse multimodal information prematurely, which makes them difficult to interpret with current computational neuroscience technologies. Therefore, in this work, we adopt the dual-stream architecture to encode the data from the two (visual and lingual) modalities with separate visual encoder $F^v$ and lingual encoder $F^l$. Motivated by the fact that human tends to simultaneously receive information from multiple modalities in learning and memory, the learning objective of our BriVL is thus to align the visual and lingual information in a unified embedding space.
Specifically, we apply the InfoNCE loss~\cite{oord2018representation} as our pre-training objective, aiming to maximize the cosine similarity of the image and text embeddings of each ground-truth pair and also minimize the cosine similarity of negative image-text pairs.

Formally, given an image $V_i$, we adopt the well-known Vision Transformer (ViT)~\cite{dosovitskiy2021an} as our visual encoder to extract image patch features $\mathbf{V}_i = [\mathbf{v}_{cls}; \mathbf{v}_1; \cdots; \mathbf{v}_{k_v-1}] \in \mathbb{R}^{k_v \times D_v}$, where $\mathbf{v}_{cls}$ denotes the [CLS] token embedding, $k_v$ denotes the patch sequence length, and $D_v$ denotes the dimension of the patch embeddings. We utilize a fully-connected layer to project the [CLS] token embedding into the final image embedding $\mathbf{f}_i^v = F^v(V_i) \in \mathbb{R}^{D}$, where $F^v$ is the visual encoder and $D$ is the dimension of the final embedding. Similarly, given a text $L_i$, we apply a pre-trained Transformer encoder BERT~\cite{devlin2018bert} as our lingual encoder to extract text sequence features $\mathbf{L}_i = [\mathbf{l}_{cls}; \mathbf{l}_1; \cdots; \mathbf{l}_{k_l-1}] \in \mathbb{R}^{k_l \times D_l}$, where $\mathbf{l}_{cls}$ denotes the [CLS] token embedding, $k_l$ denotes the text token sequence length, and $D_l$ denotes the dimension of the token embeddings. We average the output vectors of all text tokens and utilize a fully-connected layer to project the average vector into the final text embedding $\mathbf{f}_i^l= F^l(L_i) \in \mathbb{R}^{D}$, where $F^l$ is the lingual encoder.

Then we apply the InfoNCE loss for image-text semantic alignment in the final embedding space. However, with limited computational resource, the mini-batch size for large-scale pre-training tends to be small, which brings harm to contrastive learning if negative samples only come from each mini-batch.
Inspired by MoCo~\cite{he2020momentum}, we introduce the momentum mechanism to maintain massive negative samples in memory queues for contrastive learning.

\textbf{Memory queues} are used for maintaining massive negative samples. This allows us to reuse the features from preceding mini-batches. In each training iteration, the current mini-batch is enqueued to a memory queue, and the oldest mini-batch in the queue is removed. For our BriVL, we maintain a visual memory queue $\mathcal{Q}^v = \{\mathbf{\hat{q}}^v_j\}_{j=1}^{N_m}$ and a lingual memory queue $\mathcal{Q}^l = \{\mathbf{\hat{q}}^l_j\}_{j=1}^{N_m}$ to respectively store image and text features, where $\mathbf{\hat{q}}^v_j$/$\mathbf{\hat{q}}^l_j$ denotes the $j$-th stored image/text embedding and $N_m$ denotes the memory queue size. 

\textbf{Momentum encoders} are then needed to maintain the embedding consistency in the memory queues, as suggested in MoCo. We thus utilize two additional encoders: the visual momentum encoder $\hat{F}^v$ (with parameters $\hat{\theta}^v$) and the lingual momentum encoder $\hat{F}^l$ (with parameters $\hat{\theta}^l$).
The parameters of momentum encoders are updated by:
\begin{equation}
\hat{\theta}^v = m \cdot \hat{\theta}^v + (1 - m) \cdot \theta^v, 
\end{equation}
\begin{equation}
\hat{\theta}^l = m \cdot \hat{\theta}^l + (1 - m) \cdot \theta^l,
\end{equation}
where $m$ is the momentum coefficient hyper-parameter, and $\theta^v$/$\theta^l$ is the visual/lingual encoder's set of parameters.

For each image $V_i$ in mini-batch $\mathcal{B}$, we regard its paired text $L_i$ as the positive sample and all texts in the text memory queue $\mathcal{Q}^l$ as negative ones. Then the image-to-text contrastive loss is defined by the InfoNCE loss~\cite{oord2018representation}:
\begin{equation}
\mathcal{L}_\text{I2T} = -\frac{1}{N_b} \sum_{(V_i, L_i) \in \mathcal{B}} \log \frac{\text{pos}(\mathbf{f}^v_i, \mathbf{\hat{f}}^l_i, \tau)}{\text{pos}(\mathbf{f}^v_i, \mathbf{\hat{f}}^l_i, \tau) + \text{neg}(\mathbf{f}^v_i, \mathcal{Q}^l, \tau)},
\label{eq:cl_v2t}
\end{equation}
where $\mathbf{f}^v_i = F^v(V_i)$, $\mathbf{\hat{f}}^l_i = \hat{F}^l(L_i)$, $\tau$ is the temperature hyper-parameter, $N_b = |\mathcal{B}|$ is the mini-batch size, and
\begin{equation}
\text{pos}(\mathbf{f}^v_i, \mathbf{\hat{f}}^l_i, \tau)  = \exp(\cos(\mathbf{f}^v_i, \mathbf{\hat{f}}^l_i) / \tau),
\end{equation}
\begin{equation}
\text{neg}(\mathbf{f}^v_i, \mathcal{Q}^l, \tau)   = \sum_{\mathbf{\hat{q}}^l_j \in \mathcal{Q}^l} \exp(\cos(\mathbf{f}^v_i, \mathbf{\hat{q}}^l_j) / \tau).
\end{equation}
Note that the similarity of two feature vectors is measured by the cosine similarity $\cos(\cdot,\cdot)$. Similarly, for each text $L_i$ in mini-batch $\mathcal{B}$, the text-to-image contrastive loss is formulated as:
\begin{equation}
\mathcal{L}_\text{T2I} = -\frac{1}{N_b} \sum_{(V_i, L_i) \in \mathcal{B}} \log \frac{\text{pos}(\mathbf{f}^l_i, \mathbf{\hat{f}}^v_i, \tau)}{\text{pos}(\mathbf{f}^l_i, \mathbf{\hat{f}}^v_i, \tau) + \text{neg}(\mathbf{f}^l_i, \mathcal{Q}^v, \tau)},
\end{equation}
where $\mathbf{f}^l_i = F^l(L_i)$ and $\mathbf{\hat{f}}^v_i = \hat{F}^v(V_i)$. 
Finally, the pre-training objective function for BriVL is defined as:
\begin{equation}
\mathcal{L}_{\text{BriVL}} = \mathcal{L}_{\text{I2T}} + \mathcal{L}_{\text{T2I}}.
\end{equation}

\paragraph{Implementation details.}
All experiments are conducted with PyTorch on the latest distributed-training framework DeepSpeed~\cite{rajbhandari2021deepspeed}.
We adopt Randaugment~\cite{CubukZS020} (including random crops, horizontal flips, Gaussian blur, graying, color jittering, etc) for data augmentation over input images (of the size 384$\times$384).
We empirically set the initial learning rate to 1e-5 and adopt the Adam optimizer with a weight decay of 0.02 for 10 epochs. 
It takes around 120 hours to train BriVL with 16 Nvidia V100 GPUs.
We also set the other hyper-parameters uniformly as: mini-batch size $N_b = 512$, momentum hyper-parameter $m = 0.99$, temperature $\tau = 0.07$, and queue size $N_Q = 9,600$. We choose the pre-trained BERT-Base as the lingual encoder and the ViT-Base~\cite{dosovitskiy2021an} pre-trained on ImageNet as the visual encoder.
For the video modality, we sample 16 frames per video: each video is equally split into 16 segments and one frame is randomly sampled from each segment.
A single fully-connected layer is finally used to project the text/image/video embeddings to the cross-modal embedding space (with $D = 256$).

\paragraph{Differences from original BriVL.}
Although our foundation model developed for brain-inspired analysis in this paper (denoted as BriVL-Brain for convenience only) adopts almost the same framework as our original BriVL proposed in~\cite{fei2022towards}, they have three technical differences: (1) For easy model interpretability as well as fair comparison with single modality backbones (e.g., ViT and BERT), BriVL-Brain adopts pure Transformer-based architecture instead of CNN+Transformer hybrid architecture, where CNN denotes convolutional neural network. (2) As only English neural encoding datasets are available, BriVL-Brain is pre-trained with 14M English image-text pairs, while original BriVL is pre-trained with 650M Chinese image-text pairs. (3) The explainability of BriVL-Brain is explored with the help of computational neuroscience, while original BriVL is explained only in the AI research field itself.

\paragraph{Cross-modal retrieval.}
Cross-modal retrieval is the most fundamental multimodal task, which aims to retrieve proper matched samples from one modality when a query is given in the other modality. 
Note that our BriVL has been trained to measure the distance of each image-text pair by calculating the cosine similarity of image and text embeddings.
We thus simply calculate the similarities among the query and the candidate samples, and then choose the one with the highest similarity as the prediction.
Since our BriVL is trained on pure image-text datasets, we tweak the pipeline slightly to obtain video embeddings. That is, we sample multiple frames and use the average of the these frame embeddings as the video embedding for text-to-video retrieval.

\paragraph{Image-to-text generation.}
Image-to-text generation (i.e., image captioning) aims to generate descriptive sentences according to given images. For either BriVL or CLIP, we first use its visual encoder to extract a sequence of image patch features. The patch features are then transformed into a sequence of word features using an additional Transformer decoder. Finally, we adopt a fully-connected layer to predict each word.

\paragraph{Text-to-image generation.}
For text-to-image generation, we adopt the latest VQGAN~\cite{esser2020vqgan}, which consists of a codebook and a generator. Specifically, we first use BriVL's lingual encoder to extract the embedding of a given piece of text. A learnable input is randomly initialized and inputted into the generator to generate an initial image. Then we use BriVL's visual encoder to extract the embedding of the generated image and force the image embedding to be aligned with the text embedding by back-propagation (i.e., only updating the learnable input but with both BriVL and VQGAN fixed). We iterate this process multiple times to get the final generated image.

\paragraph{Neural datasets.}
The visual neural dataset used in this paper is from \cite{huth2012continuous, huthgallant}. This dataset contains Blood Oxygenation Level-Dependent (BOLD) fMRI responses from five human subjects when viewing dynamic natural movie clips. The BOLD signals were collected by a 3T Siemens Trio scanner with TR = 2s, voxel size $=2.24\times 2.24 \times 4.1$mm. The training data are collected by 12 separate 10min scans while the test data are collected by 9 separate 10min scans. Every clip of stimuli in the test set is repeatedly viewed by each subject for ten times. To increase the signal-to-noise ratio (SNR), we average the test responses across different repetitions. Finally, the training set used to fit the neural encoding models has 3,600 samples and the test set has 270 samples. The fMRI data processing includes motion correction, automatic alignment, and z-normalization. Detailed information on data acquisition and processing of this dataset can be found in \cite{huth2012continuous}. We use the publicly available processed data, without any extra processing step.

The lingual neural dataset is from \cite{pereira2018toward}. The original dataset collects fMRI signals from subjects with lingual stimuli under three different experiments. Experiment 1 shows the subjects with 180-word stimuli. Experiment 2 shows the subjects with 384 sentences from 96 passages. Experiment 3 shows the subjects with 243 sentences from 72 passages. Each passage in Experiment 2 and Experiment 3 is Wikipedia-style and is composed of several sentences about one concept varying from 24 different broad topics. There were no overlapping topics between the two experiments. 
Subjects were asked to attentively read the sentences when viewing these stimuli.
The BOLD signals of eight subjects were collected by a 3T Siemens Trio scanner with TR = 2s, and voxel size $=2.1\times 2.1\times 2.1$mm. The data preprocessing includes slice timing correction, motion correction, bias field inhomogeneity, and high-pass filtering. Detailed information on data acquisition and processing of this dataset can be found in \cite{pereira2018toward}. We directly use the publicly available processed data.

\paragraph{Neural encoding methodology.}
The classic neural encoding algorithm builds mappings between three different spaces, i.e., pixel space, feature space, and voxel space \cite{naselaris2011encoding}. First, a nonlinear transformation builds a connection between the pixel space and the feature space. Second, a linear mapping builds a connection between the feature space and the voxel space. In this paper, deep neural networks (DNNs) with powerful representation capabilities play the role of the nonlinear transformation. To compare the neural encoding performance of different DNNs and to probe into the modality preference of voxels in different ROIs, we adopt joint encoding models which can take features from multiple DNNs as input to be the linear mapping.
We did not train models based on features of each DNN and compare the results, because previous studies indicated that such an evaluation has a bias on models with large total variance but not with high unique explained variance \cite{nunez2019voxelwise}.
Specifically, we utilize the banded ridge regression proposed by \cite{nunez2019voxelwise}. Unlike other joint encoding methods, the banded ridge regression allows specific regularization for different feature spaces. Let $\mathbf{X}_1\in \mathbb{R}^{N_s \times D_1}$ and $\mathbf{X}_2\in \mathbb{R}^{N_s \times D_2}$ denote two distinct features of the stimuli, where $N_s$ is the number of stimuli, and $D_1$, $D_2$ are the feature dimensions. Let $\mathbf{Y}\in \mathbb{R}^{N_s \times N_v}$ denote the corresponding neural signals, where $N_v$ is the number of voxels. The objective of the banded ridge regression can be formulated as 
\begin{equation}
\min \mathcal{L}(\mathbf{W}_1, \mathbf{W}_2) = \left\|\mathbf{Y} -
\begin{bmatrix}
  \mathbf{X}_1 & \mathbf{X}_2
\end{bmatrix}
\begin{bmatrix}
  \mathbf{W}_1\\
  \mathbf{W}_2
\end{bmatrix} \right\|_2^2
+ \|\lambda_1 \mathbf{W}_1\|_2^2 + \|\lambda_2 \mathbf{W}_2\|_2^2.
\end{equation}
The trade-off parameters $\lambda_1$, $\lambda_2$ control the degree of regularizations. When $\lambda_1=\lambda_2$, the banded ridge regression degenerates to ridge regression. In the visual neural encoding experiments, $\mathbf{X}_1$ is the matrix of features extracted by our BriVL's visual encoder $F^v$ and $\mathbf{X}_2$ is the matrix of features extracted by the unimodally-trained ViT. In the lingual neural encoding experiments, $\mathbf{X}_1$ is the matrix of features extracted by our BriVL's lingual encoder $F^l$ and $\mathbf{X}_2$ is the matrix of features extracted by the unimodally-trained BERT. After model fitting, we can partition the encoding accuracy into the contribution of each type of features~\cite{la2022feature}. 
The coefficient of determination ($R^2$) is used as the metric of the encoding accuracy. As shown in~\cite{la2022feature}, the split $R^2$ of $\mathbf{X}_1$ and $\mathbf{X}_2$ is calculated as 
\begin{equation}
R^2_j = \frac{ \hat{\mathbf{y}}_j^T (2 \mathbf{y} - \hat{\mathbf{y}}) }{ \mathbf{y}^T \mathbf{y} },~j\in\{1,2\}
\end{equation}
where $\mathbf{y} \in \mathbb{R}^{N_s}$ is the neural response vector of one specific voxel, $\hat{\mathbf{y}}$ is the prediction of the joint encoding model, and $\hat{\mathbf{y}}_j$ is the disentangled prediction of the $j$-th feature type.
Finally, the model feature with greater explanatory power will obtain a higher $R^2$, and the sum of the $R^2$ of each feature type is equal to the $R^2$ of the banded ridge regression model.

\section*{Acknowledgements}
\textbf{Funding:}
Z.L. acknowledges National Natural Science Foundation of China (61976220). J.R.W. acknowledges National Natural Science Foundation of China (61832017), Beijing Outstanding Young Scientist Program (BJJWZYJH012 \\
019100020098), and Large-Scale Pre-Training Program 468 of Beijing Academy of Artificial Intelligence (BAAI). N.F. acknowledges the Outstanding Innovative Talents Cultivation Funded Programs 2021 of Renmin Univertity of China.
\textbf{Author contributions:}
Z.L. contributed to the original idea, model design, and experimental analysis. H.L., Q.Z., N.F. and Z.L. wrote the majority of the manuscript. H.L., Q.Z. and N.F. contributed the source code. The experiments were conducted by H.L., Q.Z., N.F. and J.W. The review and editing of the manuscript were carried out by M.D., C.D., X.Z., H.S., H.H. and J.R.W. Correspondence to Zhiwu Lu (\url{luzhiwu@ruc.edu.cn}), Huiguang He (\url{huiguang.he@ia.ac.cn}) or Ji-Rong Wen (\url{jrwen@ruc.edu.cn}).
\textbf{Competing interests:}
The authors declare no competing interests.
\textbf{Data and materials availability:}
The availability of all datasets is detailed as follows: 
(1) Six pre-training datasets:  Conceptual Captions 12M (\url{https://github.com/google-research-datasets/conceptual-12m}), Conceptual Captions 3M (\url{https://ai.google.com/research/ConceptualCaptions/}),
SBU (\url{https://www.cs.virginia.edu/~vicente/sbucaptions/}), MSCOCO  (\url{https://cocodataset.org/}), VG (\url{https://visualgenome.org/}) and Flickr30k (\url{https://docs.activeloop.ai/datasets/flickr30k-dataset}). 
(2) Two cross-modal retrieval datasets: Flickr30k (\url{https://docs.activeloop.ai/datasets/flickr30k-dataset}) and 
MSR-VTT (\url{https://www.microsoft.com/en-us/research/publication/msr-vtt-a-large-video-description-dataset-for-bridging-video-and-language/}). 
(3) Image-to-text generation dataset: MSCOCO (\url{https://cocodataset.org/}). 
(4) Neural datasets: the visual neural dataset ``shortclips''  (\url{https://gin.g-node.org/gallantlab/shortclips}), and the lingual neural dataset (\url{https://evlab.mit.edu/sites/default/files/documents/index2.html}).
The pre-trained network and testing code are available at \url{https://github.com/RERV/BriVL-Brain} under a Creative Commons Attribution-Non Commercial-No Derivatives 4.0 International Licence (CC BY-NC-ND) license.

\bibliographystyle{naturemag}
\bibliography{sample}

\end{document}